\documentclass[sigconf]{acmart}

\AtBeginDocument{%
  }

\usepackage{latexsym}
\usepackage{graphicx}
\usepackage{booktabs} %
\usepackage{color}  %
\usepackage{subcaption}
\usepackage{caption}
\usepackage{colortbl} %
\usepackage{arydshln}
\usepackage{mathtools}
\usepackage{multirow}
\usepackage{framed}
\usepackage{hyperref}
\usepackage{tablefootnote}
\usepackage{makecell}
\usepackage{colortbl}
\usepackage{xcolor}
\usepackage{siunitx}
\usepackage{stfloats}
\usepackage[inline]{enumitem}

\newcommand{\squishlist}{
 \begin{list}{$\bullet$}
  { \setlength{\itemsep}{0pt}
     \setlength{\parsep}{1pt}
     \setlength{\topsep}{1pt}
     \setlength{\partopsep}{0pt}
     \setlength{\leftmargin}{1.5em}
     \setlength{\labelwidth}{1em}
     \setlength{\labelsep}{0.5em} } }
 \newcommand{\squishend}{\end{list}}

\setcopyright{rightsretained}
\copyrightyear{2025}
\acmYear{2025}
\acmDOI{XXXXXXX.XXXXXXX}

\acmConference[WWW '25]{Make sure to enter the correct conference title from your rights confirmation email}{June 03--05,
  2018}{Woodstock, NY}
\acmISBN{978-1-4503-XXXX-X/18/06}

\begin{document}

\title{\textsc{Neon}: News Entity-Interaction Extraction for Enhanced Question Answering}

\author{Sneha Singhania}
\authornote{Work done while interning at Microsoft Research}
\affiliation{%
  \institution{Max Planck Institute for Informatics}
  \city{Saarbrücken}
  \country{Germany}}
\email{ssinghan@mpi-inf.mpg.de}

\author{Silviu Cucerzan}
\affiliation{%
  \institution{Microsoft Research}
  \city{Redmond}
  \country{USA}}
\email{silviu@microsoft.com}

\author{Allen Herring}
\affiliation{%
  \institution{Microsoft Research}
  \city{Redmond}
  \country{USA}}
\email{allenh@microsoft.com}

\author{Sujay Kumar Jauhar}
\affiliation{%
  \institution{Microsoft Research}
  \city{Redmond}
  \country{USA}}
\email{sjauhar@microsoft.com}

\renewcommand{\shortauthors}{Singhania et al.}

\begin{abstract}

Capturing fresh information in near real-time and using it to augment existing large language models (LLMs) is essential to generate up-to-date, grounded, and reliable output. This problem becomes particularly challenging when LLMs are used for informational tasks in rapidly evolving fields, such as Web search related to recent or unfolding events involving entities, where generating temporally relevant responses requires access to up-to-the hour news sources.
However, the information modeled by the parametric memory of LLMs is often outdated, and Web results from prototypical retrieval systems may fail to capture the latest relevant information and struggle to handle conflicting reports in evolving news. To address this challenge, we present the \textsc{Neon} framework, designed to extract emerging entity interactions---such as events or activities---as described in news articles. \textsc{Neon} constructs an entity-centric timestamped knowledge graph that captures such interactions, thereby facilitating enhanced QA capabilities related to news events. Our framework innovates by integrating open Information Extraction (openIE) style tuples into LLMs to enable in-context retrieval-augmented generation. This integration demonstrates substantial improvements in QA performance when tackling temporal, entity-centric search queries. Through \textsc{Neon}, LLMs can deliver more accurate, reliable, and up-to-date  responses. 

\end{abstract}




\maketitle

\section{Introduction}

\begin{figure*}[t!]
    \centering
     \includegraphics[width=0.9\linewidth]{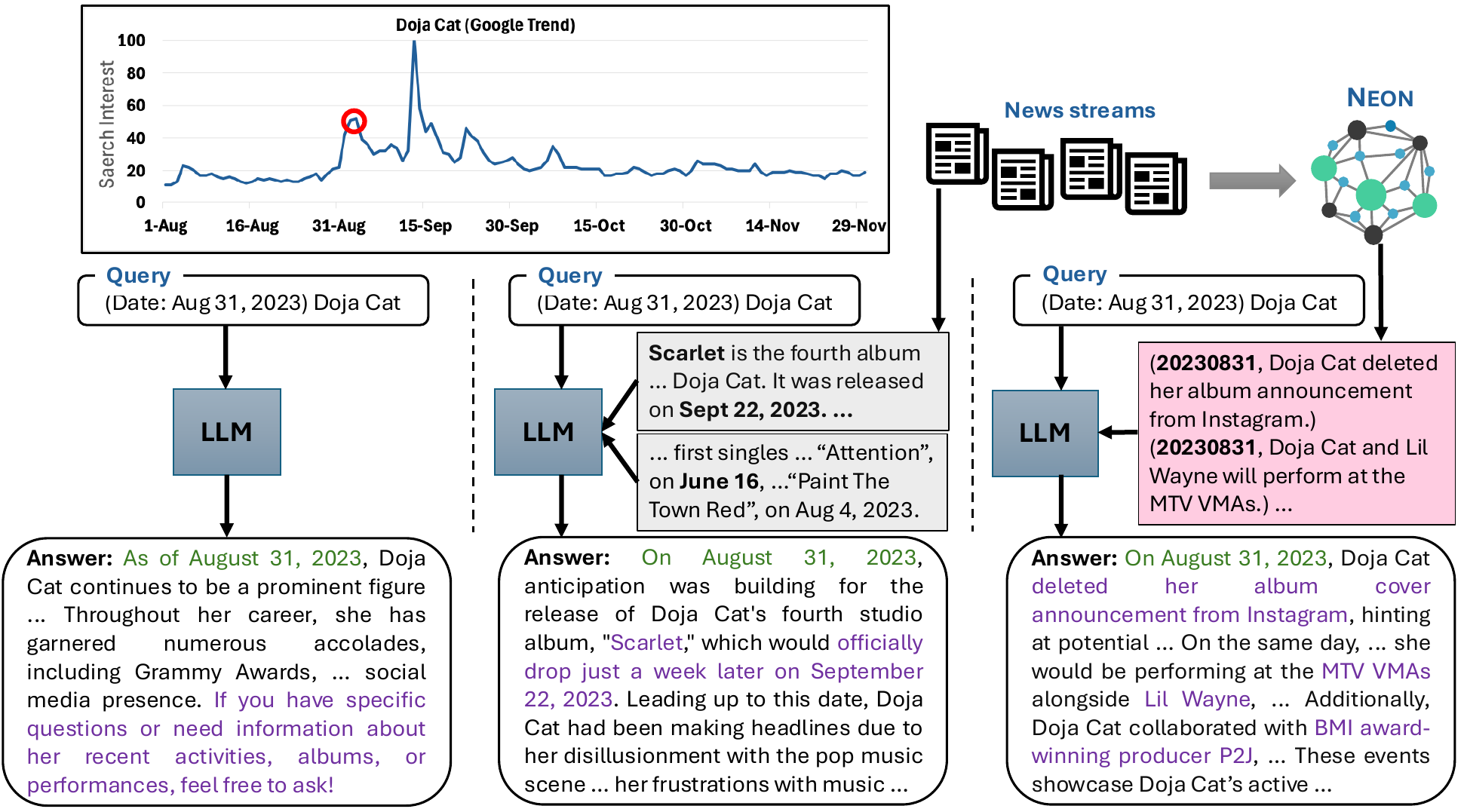}
     \Description{Example for entity-centric time-specific QA.}
     \caption{Example for entity-centric, time-specific QA. Graph shows search interest for \textit{Doja Cat} over a four-month period. The bottom part illustrates response generation at one of the peaks (\texttt{31 August}) using three different techniques: (i) zero-shot prompting, (ii) news snippets based prompting, (iii) augmenting tuples from our \textsc{Neon} graph for enhanced answer generation.}
     \label{fig:overview}
\end{figure*}

In recent years, large language models ~\cite{peters-etal-2018-deep, devlin-etal-2019-bert}, pretrained on extensive web-scale datasets, have gained widespread usage as human task assistants, offering expert-level knowledge learned during pretraining~\cite{OpenAI2023GPT4TR}. However, given the dynamic nature of information in the real world, incorporating temporal awareness into these AI systems has become increasingly important. The ability to inject up-to-date information is essential for effectively addressing user needs in a fast evolving world. Despite the notable capabilities of these systems, research on downstream tasks involving temporal aspects has uncovered several limitations. Key issues include temporal misalignment~\cite{luu-etal-2022-time, jang-etal-2022-temporalwiki}, lack of temporal generalization~\cite{jin-etal-2022-lifelong}, limited temporal reasoning capabilities~\cite{xiong-etal-2024-large}, and disproportionate handling of implicit versus explicit temporal conditions when answering knowledge-intensive questions~\cite{10.5555/3666122.3668252}.

These challenges are particularly acute in the news domain, where new content is generated rapidly, and producing reliable, grounded responses cannot solely depend on static, pretrained knowledge. Several studies have demonstrated that LLMs can perform well on domain-specific question answering tasks~\citep[][\textit{inter alia}]{NEURIPS2020_1457c0d6, kaddour2023}, including on popular news datasets. More recent approaches leverage LLMs through the retrieval-augmented generation (RAG) paradigm~\cite{realm, NEURIPS2020_6b493230, ram-etal-2023-context} to address some of the aforementioned limitations. By using trained retrievers to fetch relevant supporting documents, these methods aim to enhance the accuracy of generated responses. However, there has been relatively limited focus~\citep{ning-etal-2020-torque, si2023prompting, 10.5555/3666122.3668252} on systematically retrieving temporal text snippets for answering news-related questions.

In this work, we focus on generating reliable and temporally grounded responses to user queries, with a particular emphasis on entity-centric queries. Our choice is motivated by several factors: (i) users frequently inquire about entities during real-world events~\cite{10.1145/1571941.1571989, 10.1145/2187836.2187916}, (ii) entity-centric information tuples of the form \textit{(subject, predicate, object)} are highly valuable in Knowledge Base Question Answering (KBQA) tasks~\cite{10.1145/3459637.3482416, tempoqr}, and (iii) user queries are often vague, ambiguous, and telegraphic, with implicit intent tied to the mentioned entities~\cite{Balog2018EntityOrientedS, 10.1145/2488388.2488484}. For example, Figure~\ref{fig:overview} shows the Google trend line (\href{https://trends.google.com/trends/explore?cat=3&date=2023-08-01\%202023-11-30&q=Doja\%20Cat}{https://trends.google.com/trends})
for the popular entity \textit{Doja Cat} during the \texttt{August-November, 2023} time frame, exhibiting peak search interest on specific dates, particularly around major events, e.g., on \texttt{August 31} and \texttt{September 15}.

When a query such as ``Doja Cat'' or ``Doja Cat news'' is given in the prompt to an LLM such as GPT-4~\cite{OpenAI2023GPT4TR}, the model likely generates an entity-centric response based on sources like Wikipedia, one of the largest resources in LLM pretraining datasets~\cite{JMLR:v21:20-074, together2023redpajama}. Although Wikipedia provides valuable overviews and highlights about key events, it often lacks the granular, timely updates required to answer entity-centric queries involving recent developments. Even if the user query is reformulated to specify a date---e.g., ``(Date: August 31, 2023) Doja Cat'' ---LLMs often struggle to produce precise, up-to-date response, in particular when the date in question is beyond the LLM's knowledge cut-off and the system must defer to external sources through RAG mitigation.

This limitation is illustrated in Figure~\ref{fig:overview}, where we selected a random peak in the search interest plot. Without additional information sources and/or additional context, GPT-4 provides a generic response to the prompt ``(Date: August 31, 2023) Doja Cat''. However, we observe improvements in temporal relevance and accuracy through RAG mitigation when supporting snippets retrieved from news streams are incorporated into the LLM's prompt. Here, LLM still needs to reason and extract relevant information across these multiple snippets.
As \citet{liu-etal-2024-lost} note, both the size of the context and the positioning of relevant information within the context significantly impact model performance. Instead of directly augmenting chunks of information,
we propose \textsc{Neon} (\underline{\textbf{N}}ews \underline{\textbf{E}}ntity Interacti\underline{\textbf{ON}}s) that captures interactions such as events and activities between entities in a more compact form (single sentences or propositions) and use these for enhanced response generation.

\textsc{Neon} is an entity-centric, time-stamped graph constructed from a rich stream of news sources. Unlike prior information extraction (IE) methods that leverage LLMs, we identify, disambiguate, and label the entities from news streams first, and then employ the entity markup information in LLM prompts to generate open information extraction (openIE) style semi-structured news entity-interactions. This method helps us achieve higher recall by avoiding the bottleneck of canonicalization. Once \textsc{Neon} is constructed, we use various temporal indexing strategies to effectively retrieve relevant information from \textsc{Neon} for answering temporal user queries. For instance, in Figure~\ref{fig:overview}, two \textsc{Neon} entity-interactions related to \textit{Doja Cat} are augmented for response generation.

We test the effectiveness of \textsc{Neon} by collecting 3,000 real-world queries from the Bing search engine on specific dates for 50 sampled entities. These entities were chosen with various considerations in mind, including demographic diversity, the balance between popular and long-tail entities, and a mix of person and organization-type entities, as explained further in Section~\ref{sec:experimental_setup}. To perform temporal QA generation, we augment the LLM prompts with retrieved \textsc{Neon} tuples (illustrative examples provided in Table~\ref{tab:neon_exmaples}). Since answers to temporal queries can be subjective, and LLMs have proven to be effective evaluators~\cite{liu-etal-2023-g, wang2024helpsteer2opensourcedatasettraining}, particularly on downstream tasks within the news domain~\cite{xu-etal-2023-instructscore, chiang-lee-2023-large}, we employ LLMs to automatically assess the quality of responses on a 3-point Likert scale. We also corroborate the automatic evaluations through human assessment. Our experiments reveal that integrating \textsc{Neon} entity-interactions into LLM prompts improves the temporal relevance and overall quality of the answers.

\noindent
The salient contributions of this work are:
\squishlist
\item[(i)] We formulate the problem of generating enhanced responses for temporal entity-centric queries.
\item[(ii)] We present a novel approach for extracting openIE style entity-interactions tuples from news streams.
\item[(iii)] We evaluate the generated answers using both automated LLM-based scoring and human evaluations on a diverse set of real-world queries, demonstrating significant improvements in response quality when incorporating \textsc{Neon}.
\squishend

\section{\textsc{Neon}}

In this section, we present \textsc{Neon}, an openIE-style knowledge graph for modeling entity-interactions from news streams. We outline the graph construction pipeline, describe optimized indexing techniques for efficient temporal retrieval, and then detail our approach for temporal question-answering, demonstrating how \textsc{Neon} enables accurate, contextually relevant responses.

\begin{table}[t!]
\resizebox{\columnwidth}{!}
{
    \centering
    \begin{tabular}{p{0.08cm}p{7.9cm}}
    \toprule
    \multirow{6}{*}{\rotatebox{90}{\textsc{Neon($\mathcal{M}_1$)}}}
  & (\textit{20230502}, \textbf{Doja Cat} made debut appearance at \textbf{Met Gala}) \\
  & (\textit{20230502}, \textbf{Doja Cat} dressed as \textbf{Choupette}) \\
  & (\textit{20230502}, \textbf{Doja Cat} was styled by \textbf{Brett Alan Nelson})\\
  & (\textit{20230831}, \textbf{Doja Cat} will perform at the 2023 \textbf{MTV VMAs})\\
  & (\textit{20230831}, \textbf{Doja Cat's} announcement debuting the cover art for ``Scarlet'' was removed from \textbf{Instagram})\\
     \midrule
     \multirow{5}{*}{\rotatebox{90}{\textsc{Neon($\mathcal{M}_2$)}}}
  & (\textit{20230502}, \textbf{Doja Cat} and \textbf{Jared Leto} paid homage to Lagerfeld) \\
  & (\textit{20230531}, \textbf{Doja Cat} and \textbf{Demi Lovato} will perform at VMAs)\\
  & (\textit{20230831}, \textbf{Doja Cat} collaborates with Afrobeats producer \textbf{P2J})\\
  & (\textit{20230831}, \textbf{Doja Cat} posted on \textbf{Twitter})\\
  & (\textit{20230831}, \textbf{Doja Cat} deleted a post on \textbf{Twitter})\\
    \bottomrule
    \end{tabular}}
    \caption{Samples from \textsc{Neon} variants. The subject and object entities are bold-faced.}
    \label{tab:neon_exmaples}
\end{table}

\subsection{Graph Construction}\label{sec:neon}

\textsc{Neon} is a knowledge graph wherein nodes are entities and edges capture interactions (events or activities) between these entities. Formally, \textsc{Neon} is defined as a graph $\mathcal{G} = (E, T, R)$, where $E$ is the set of entities, $I$ is the lexicalized interaction descriptions or propositions in natural language, $T$ is the time frame, and $R$ refers to timestamped entity-interactions related both subject ($s$) and object ($o$) entities, defined as $R = (t,\texttt{NEI}(s, o))$, with $s, o \in E$ and $t \in T$. 

To construct \textsc{Neon}, we process a large set of news streams from various internet sources. Each article is time-stamped according to its publication date, typically extracted from the source URL. The construction pipeline follows these steps:
\squishlist
    \item[1.] \textbf{Entity identification:}  Each news story is preprocessed to extract only the main content, which is analyzed using an entity linking tool called NEMO~\cite{nemo}. This step annotates all named entity mentions in text, including co-references and additional information like dates and addresses, with a simple XML markup.
    \item[2.] \textbf{Sentence segmentation:} The article content is segmented into sentences using a proprietary tool, similar to spaCy\footnote{\url{https://spacy.io/api/sentencizer}}, and entities identified in previous step are tracked within each sentence. 
    \item[3.] \textbf{Text chunking:} The article’s main content is then split into overlapping chunks ($C$) of $m$ sentences (where $m$ is a hyperparameter). For indexing, each chunk ($c \in C$) is represented as $c = (X_{1...m}, E', t)$, where $X_{1...m}$ denotes a set of $m$ sentences, $E'$ is the set of named entities identified within $X_{1...m}$ and $E' \subset E$, and $t$ is the article's timestamp.
    \item[4.] \textbf{Chunk Deduplication:} To manage paraphrased information across different chunks, near-duplicate chunks are filtered out using trigram representations and Jaccard similarity.
\squishend

Once the set of news chunks is obtained, next we explain how information extraction (IE) techniques that employ LLMs are applied to each chunk to construct the \textsc{Neon} graph.

\paragraph{\textbf{Overview}} 
We design task-specific prompts for the LLM that explore two IE variants for graph construction.
In the first variant, termed \textsc{Neon}($\mathcal{M}_1$), LLM generates news entity-interactions $R$ using only the subject $s$ and the corresponding news chunk $c$ that mentions $s$. In the second variant, $R$ is generated by incorporating both the subject $s$ and object $o$ entities, along with the news chunk $c$ that contains mentions of both ($s$, $o$).

The first variant focuses on entity-interactions centered around a single subject entity, making it ideal for cases where the entity (e.g., a person or organization) is the primary focus of the news chunk. The second variant, which includes both subject and object entities, enables the model to infer direct relationships or events between the pair, resulting in a more detailed graph that captures complex associations and evolving scenarios more accurately (see Section~\ref{sec:results}). Both approaches support a wide range of interactions without relying on predefined relationships. Given our goal of solving temporal QA, we prompt the LLM to directly generate news entity-interactions $R$ in a lexicalized form~\cite{oguz-etal-2022-unik, jia2024faithful}, thereby enabling \textsc{Neon} to be directly augmented into the temporal QA prompt.

\paragraph{\textbf{\textsc{Neon}($\mathcal{M}_1$)}}\label{sec:neon1}
Our primary objective of the $\mathcal{M}_1$ variant is to automatically construct an entity-centric, time-stamped KB using a target set of subject entities, facilitated by LLM prompting. The model identifies neighboring nodes (object entities already marked in text) and the connecting edges (interactions or descriptions) between the subject and object entities using the news chunks included in the prompt. 
The process is formally outlined in the following steps:

\squishlist
\item[1.] \textbf{Target Subjects:} Begin with a target set of subjects $S$, s.t. $S \subset E$.
\item[2.] \textbf{Chunk Retrieval:} For each subject $s \in S$, retrieve all the news chunks $c = (X_{1...m}, E', t)$ that contain mentions of $s$, i.e., $s \in E'$.
\item[3.] \textbf{Prompt Construction:} Create a prompt by explicitly mentioning $s$ and the corresponding news chunk $c$ that employs the entity markup (``<e>'' and ``</e>'') in $X_{1...m}$ to identify the entities $E'$ in the news chunk. Employ the prompt with LLM to generate a list of valid news entity-interactions $r_s = \texttt{NEI}(s, o)$, where $o$ is drawn by the LLM from the marked up entities, $\texttt{NEI}$ is LLM's implicit verbalization function. Each news entity-interaction is assigned a timestamp, resulting in $r_s = (t, \texttt{NEI}(s, o))$, where $t$ is derived from the timestamp of $c$.
\item[4.] \textbf{Iteration:} Repeat this process for all subject-chunk pairs $(s, c)$, iteratively building the KB for each subject entity, $\mathcal{M}_1 = \bigcup\limits_{s \in S} r_s$.
\squishend

Although this approach emphasizes information related to the target entities, scaling up the set $S$ can lead to a more diverse and comprehensive KB. The chunk size ($m$ in $X_{1...m}$) can be large enough to provide detailed context around the subject entity while still fitting within the LLM's context window. This setup enables a single prompt to generate a longer list of relational tuples, with the added context potentially improving entity-interaction extraction. The prompt template for \textsc{Neon}($\mathcal{M}_1$) construction is shown in ~\ref{fig:llm_prompt1}.

\paragraph{\textbf{\textsc{Neon} ($\mathcal{M}_2$)}} \label{sec:neon2}
In contrast to $\mathcal{M}_1$, core idea behind $\mathcal{M}_2$ is to leverage the co-occurrence of entity mentions within the news chunks as potential target pairs of subject and object entities. We start with the predefined set of subject entities and identify entity co-occurrences in the news chunks, then determine candidate entity pairs based on their TF-IDF scores. Once the target pairs are obtained, the LLM is prompted to perform only the interaction extraction task. The overall process mirrors that of \textsc{Neon}($\mathcal{M}_1$), but with slightly different initializations. Specifically, the steps are as follows:

\squishlist

\item[1. ] \textbf{Target Pair Selection}: Begin with a target set of subjects $S$, s.t. $S \subset E$. Then identify co-occurring entity pairs $(s, o)$, where $s \in S, o \in E$, within all news chunks with prominent TF-IDF scores. These pairs become the target set for interaction extraction.

\item[2.] \textbf{Chunk Retrieval}: For each entity pair $(s, o)$, retrieve all relevant news chunks $c = (X_{1...m}, E', t)$, where $s, o \in E'$.

\item[3.] \textbf{Chunk Batching}: Since $(s, o)$ are predefined, the chunk context can be narrowed to focus on $(s, o)$, enabling multiple chunks to fit within a single prompt for extraction. Hence, retrieved chunks are partitioned into batches of size $k$. 

For a set of $n$ retrieved news chunks $C$ sorted based on its time-stamp, batching gives $C_B = \{C_{1...k}, \dots, C_{n-k+1...n}\}$, where each batch contains $k$ chunks. If $n$ is not divisible by $k$, the last batch will contain <$k$ chunks. This time-based sorting and batching approach allows information from similar dates---and thus similar events---to be grouped within the same prompt. 

\item[4.] \textbf{Prompt Construction}: Construct a prompt by explicitly including both the subject $s$ and object $o$, along with a chunk batch $C' \in C_B$. In each chunk within the batch, use markups (``<e>'' and ``</e>'') to highlight all entity mentions. The LLM is then prompted to generate news entity-interactions $r_{so} = \texttt{NEI}(s, o)$, where $\texttt{NEI}$ is LLM's implicit verbalization function. Since the news chunks in $C'$ may contain different time-stamps, we collect the unique time-stamps $T'$ from each chunk. The extractions are then expanded as $R_{soC'} = \bigcup\limits_{t \in T'} (t, \texttt{NEI}(s, o))$.

\item[5.] \textbf{Iteration}: 
Repeat this process for all identified $(s, o)$ target pairs and chunk batches $C' \in C_B$ to form a consolidated KB $\mathcal{M}_2 = \bigcup\limits_{(s, o)}\bigcup\limits_{C'} R_{soC'}$.

\squishend

This variant often yields more generations due to explicit $(s, o)$ pairs; however, it may occasionally cause the LLM to generate sentences involving $(s, o)$ without a valid interaction, or return an empty list if no interaction is confidently detected. Figure~\ref{fig:llm_prompt2} gives the prompt for \textsc{Neon}($\mathcal{M}_2$) construction. The chunk batch size $k$ (hyperparameter) directly affects performance: a small $k$ is cost-inefficient with more LLM calls and reduced contextual detail, while a large $k$ risks ``lost-in-the-middle''~\cite{liu-etal-2024-lost} and information overload~\cite{Singhania2024RecallTA} issues.
Table~\ref{tab:neon_exmaples} gives sample generations for both \textsc{Neon} variants. The subject is the entity \textit{Doja Cat} in both variants and the object varies for \textsc{Neon}($\mathcal{M}_2$).

\subsection{Temporal Datastore}\label{sec:temporal_datastore}

Each \textsc{Neon} tuple includes a \textit{time-stamp}, followed by the LLM-generated news entity-interaction. We represent \textsc{Neon} as a datastore ($\mathcal{D}$) with each entry as a tuple $(t_d, d)$, where $d$ is the LLM-generated news entity-interaction, and $t_d$ is the time-stamp of $d$. Following prior work~\cite{DBLP:journals/corr/abs-2201-10005, contriever}, we employ off-the-shelf models to perform dense indexing of $\mathcal{D}$, creating a temporal datastore that supports efficient retrieval. User queries $Q$ (original and reformulated) are similarly converted into dense vectors using the same model, allowing retrieval of the top-$k$ relevant entity-interactions by maximizing semantic similarity. This approach is retriever-agnostic, allowing for integration with any suitable retriever model.

To ensure temporal relevancy while retrieving content for answering each user query $q \in Q$, we reformulate the queries to explicitly mention the normalized timestamp $q' = \texttt{REFORM}(t_q, q)$, where $t_q$ is the query timestamp. We then experiment with the following two retrieval strategies:

\squishlist
\item[1.] \textbf{Temporal Retrieval}: Retrieve the top-$k$ tuples from $\mathcal{D}$ using $q$, where the tuples timestamp $t_d$ exactly matches the query timestamp $t_q$. If fewer than $k$ tuples match $t_q$, we retrieve additional tuples within a $\pm r$ day range of $t_q$ to complete the top-$k$ results, with $r$ being a hyperparameter.
\item[2.] \textbf{Generic Retrieval}: Retrieve the top-$k$ tuples solely based on semantic similarity between $q'$ and each document in $\mathcal{D}$.
\squishend

While temporal retrieval can enhance relevance for time-specific queries, it may yield zero tuples when the news coverage is sparse for query dates. In contrast, generic retrieval consistently provides tuples by first retrieving temporally relevant tuples and, if necessary, uses a back-off approach to additionally retrieve semantically relevant tuples.

\subsection{Temporal Question Answering}

\begin{figure}[t!]
     \centering
     \includegraphics[width=\linewidth]{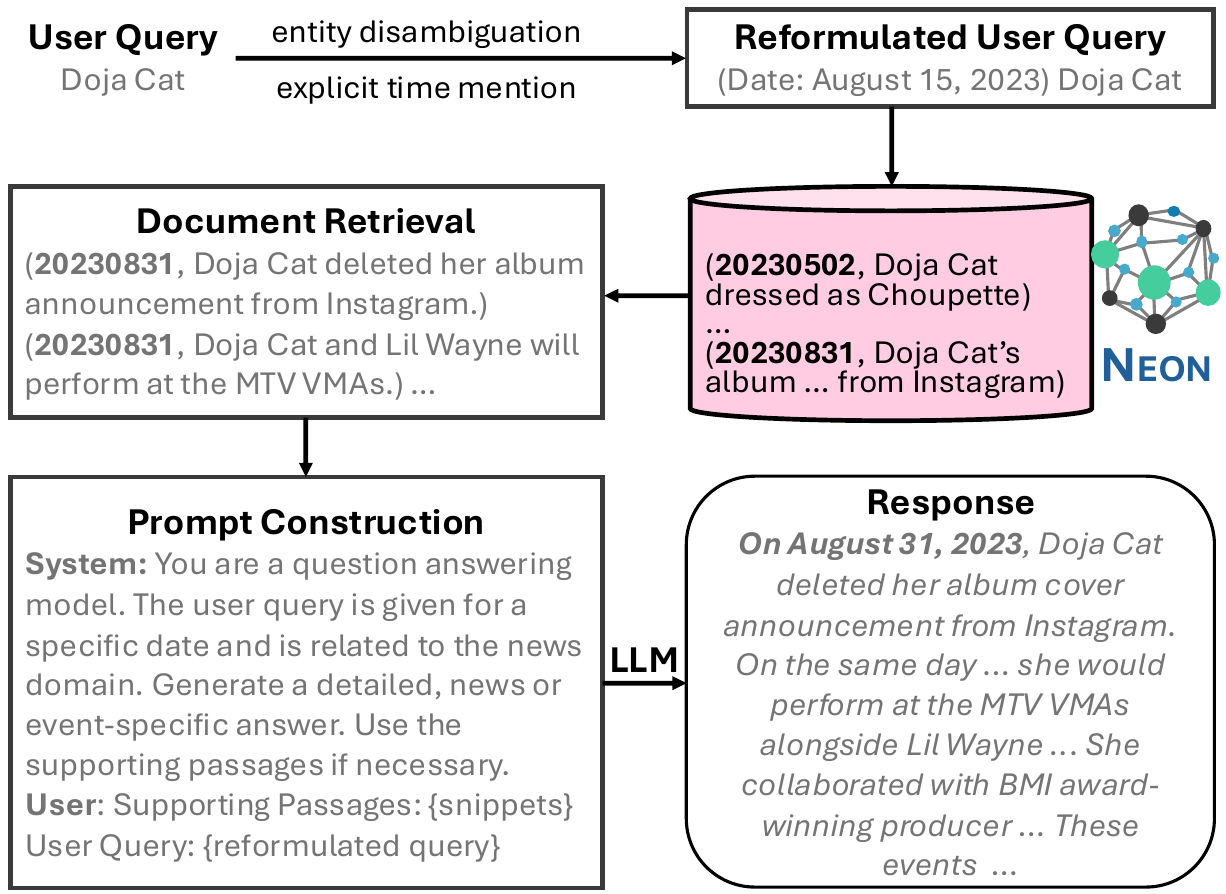}
     \Description{Temporal QA pipeline}
     \caption{Implemented temporal QA pipeline}
     \label{fig:pipeline}
\end{figure}

The final step involves generating answers for temporal queries by integrating \textsc{Neon} tuples into LLM prompts. This enables the system to deliver contextually relevant, temporally accurate, and up-to-date answers. The key steps in our approach are as follows:

\squishlist
\item[1.] User enters a (telegraphic) temporal query $q$ at timestamp $t_q$
\item[2.] $q$ is reformulated into $q'$ by performing named-entity disambiguation~\cite{cucerzan-2007-large} and explicitly mentioning $t_q$ in natural language.
\item[3.] Using the reformulated query $q'$, we apply the retrieval methods mentioned in Section~\ref{sec:temporal_datastore} to obtain top-$k$ relevant \textsc{Neon} tuples.
\item[4.] The top-$k$ \textsc{Neon} tuples are augmented into LLM prompt for RAG based generation. Figure~\ref{fig:pipeline} gives the prompt template.
\item[5.] LLM generates a response to the temporal query using the supporting information.
\squishend

Figure~\ref{fig:pipeline} provides an overview of our temporal QA pipeline, illustrating RAG mitigation for response generation and demonstrating how the \textsc{Neon} graph integrates into this system. 
Unlike conventional QA systems, our aim is to generate relevant and temporally reliable responses.

\section{Experimental Setup}\label{sec:experimental_setup}

\textbf{Entities and Queries.} We selected 50 entities by using stratified sampling across four categories---artists, companies, leaders, and pioneers---from Time's 100 most influential people~\footnote{\url{https://time.com/collection/100-most-influential-people-2023}} and Fortune500 companies~\footnote{\url{https://fortune.com/ranking/fortune500/2023/}} to ensure a balanced diversity in category, demographics, and popularity. For each entity, we collected a large volume of user queries from Bing search logs over a four-month period (\texttt{August-November, 2023}). Because we aimed to generate temporal answers for emerging events, we focused on entity-centric queries logged during \textit{spiking dates}---periods of heightened entity activity/presence in user queries.

Spiking dates were identified through short-term trends in query volume. Specifically, we calculated a 3-day rolling sum of daily query counts, flagging significant spikes as dates where the rolling sum exceeded one standard deviation above the mean. This threshold effectively identifies periods of elevated user interest, often triggered by shifts in attention or external events impacting the entity. Furthermore, to preserve user privacy, we filtered queries requested by fewer than five distinct individuals, yielding nearly 3000 temporal user queries. Table~\ref{tab:queries} provides a few examples of representative queries from each category and entity popularity.

\begin{table}[t!]
    \begin{tabular}{p{0.1cm}p{1.2cm}llp{3.8cm}}
    \toprule
        & \bfseries Entity & \bfseries Type & \bfseries Date & \bfseries Query  \\
    \toprule
    
    \multirow{6}{*}{\rotatebox{90}{Artists}} 
                    & \multirow{3}{*}{\makecell{Neil \\Gaiman}}  & \multirow{3}{*}{tail} & Sept 5  & neil gaiman award  \\
                & &  &  \multirow{2}{*}{Sept 5}  & which novel did neil gaiman coauthor with terry pratchett \\
    
                & \multirow{3}{*}{\makecell{Austin \\Butler}}  & \multirow{3}{*}{head} & Aug 17  & austin butler as elvis   \\
                &  &  &  \multirow{2}{*}{Aug 18}  & what are the latest projects of austin butler    \\
    \midrule
    
    \multirow{4}{*}{\rotatebox{90}{Companies}} 
                & \multirow{2}{*}{\makecell{Berkshire \\Hathaway}}  & \multirow{2}{*}{tail}  & Nov 6   & berkshire hathaway cash pile   \\
                & & & Nov 7   & berkshire hathaway stock \\
                & \multirow{2}{*}{\makecell{Alphabet \\ Inc.}}  & \multirow{2}{*}{head} & Aug 24  & alphabet stock price today   \\
                &  &  & Sept 14 & alphabet layoffs   \\
    \midrule
    
   \multirow{4}{*}{\rotatebox{90}{Leaders}} 
                & \multirow{2}{*}{\makecell{Sherry \\Rahman}}  & \multirow{3}{*}{tail}  & Aug 25  & climate financing by rahman \\
                & & & Oct 5  & pak minister sherry rehman \\
                & \multirow{2}{*}{\makecell{Olaf \\ Scholz}} & \multirow{2}{*}{head} & Sept 3  & olaf scholz falls  \\
                & & & Sept 4  & olaf scholz eye patch    \\
    \midrule
    
    \multirow{4}{*}{\rotatebox{90}{Pioneers}} 
                & \multirow{2}{*}{\makecell{Bella \\Hadid}}  & \multirow{2}{*}{tail}  & Oct 27  & bella hadid cancelled    \\
                & & & Oct 31  & bella hadid closing coperni    \\
                & \multirow{2}{*}{Doja Cat} & \multirow{2}{*}{head} & Aug 28  & doja cat beefs with fans \\
                & & & Aug 29  & doja cat feels free \\
    \bottomrule
    \end{tabular}
    
    \caption{User query samples from our dataset}
    \label{tab:queries}
    \vspace{-3mm}
\end{table}

\textbf{News Articles.}
To construct \textsc{Neon}, we collected news streams from 500 distinct sources, with a median of 200 articles per source and a maximum of 9,000 articles. Articles varied in length and were preprocessed to create metadata information, including disambiguated named entity mentions (e.g., persons, orgs., and locations), their surface forms, and article timestamps. The metadata helps in mapping target subject entities to their corresponding news articles. We also identified duplicate content across articles and publication dates; for temporal relevance, we retained unique content mapped to all associated sources and dates to construct a more complete \textsc{Neon} graph. Table~\ref{tab:dataset-stats} gives the final dataset statistics. The counts of head/tail entities are based on $70^{\text{th}}$ pth of the total article count.

\begin{figure}[htbp]
     \centering
     \includegraphics[width=\linewidth]{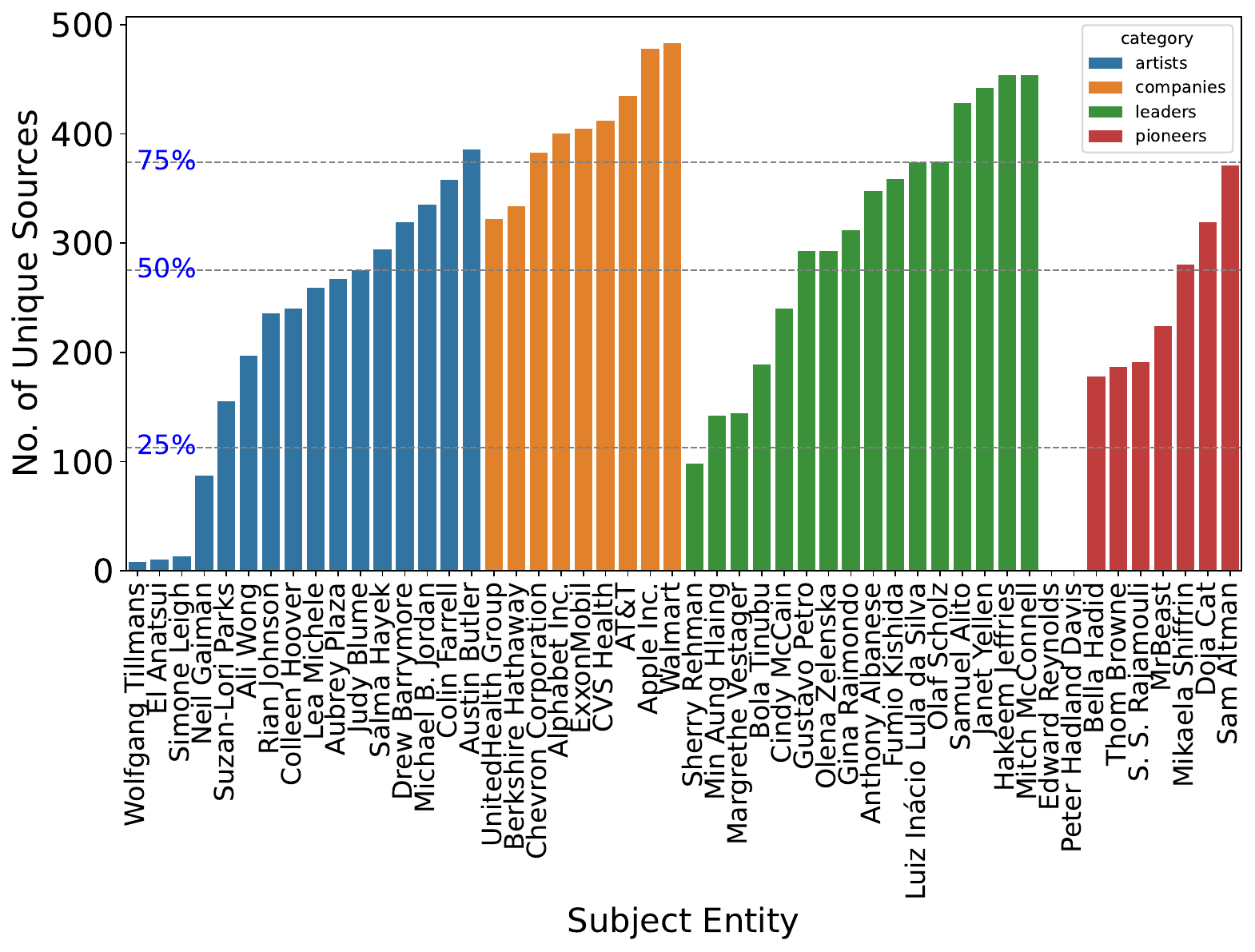}
     \Description{Entity News Domain Insights}
     \vspace*{-5mm}
     \caption{Coverage for 50 entities in 500 diverse news sources over a period of one year (2023)}
     \label{fig:entity_news}
     \vspace{-5mm}
\end{figure}

Figure~\ref{fig:entity_news} shows the count of unique news sources per subject entity, grouped by the four categories. Dashed lines at 25\%, 50\%, and 75\% on the y-axis represent quartile marks, offering a visual reference for comparing entity news coverage. Distinct patterns emerge across categories: leaders and companies dominate the news space, while artists and pioneers, though influential, receive selective coverage, with some receiving minimal or near-zero coverage.

\begin{table}[t!]
    \centering
    {
  \begin{tabular}{lc}
  
\toprule
    \textbf{Entity sources} & Time100 and Fortune500 \\
    \textbf{No. of subjects} & 50\\
    \textbf{No. of head subjects} & 15\\
    \textbf{No. of tail subjects} & 35\\
\hdashline
    \textbf{No. of news domains} & 500 \\
    \textbf{Median no. of news} & $\sim$200 \\
    \textbf{Max no. of news} & $\sim$9000 \\
\hdashline
    \textbf{No. of user queries} & 3000 \\
    \textbf{Spiking date range} & August--November 2023 \\
\bottomrule
  \end{tabular}
    }
    \caption{Dataset Statistics}
    \label{tab:dataset-stats}
    \vspace{-4mm}
\end{table}

\textbf{Evaluation.}
Since our temporal QA task involves generating subjective responses for evolving news events, traditional evaluation methods prove challenging due to the absence of an established ground-truth dataset. To address this, we employed an automated evaluation framework that leverages LLMs, which have recently gained prominence~\cite{kocmi-federmann-2023-large, wan2024tntllmtextminingscale}, even in news-related tasks~\cite{xu-etal-2023-instructscore, chiang-lee-2023-large}, as an efficient means for output assessment. Our evaluation focuses on three key attributes: (1) \textit{Helpfulness}, assessing the extent to which the response addresses the user's query by incorporating additional relevant entities---such as people, events and locations---beyond the primary subject; (2) \textit{Relevance}, ensuring that the response is accurately aligned with the context of the query, delivering information that is specific to the date or time frame referenced; (3) \textit{Faithfulness}, evaluating the extent to which the response is grounded in the supporting passages provided, relying exclusively on verifiable information from these sources to produce a trustworthy answer. Each attribute is rated on a 3-point Likert scale, in alignment with established practices in prior evaluation studies~\cite{wang2024helpsteer2opensourcedatasettraining, maddela-etal-2021-controllable}.

Notably, work by \citet{dettmers2023qlora} observed that the order and detail of attribute ratings can influence evaluation outcomes. To mitigate potential bias, we prompt the LLM separately for each attribute, tripling the number of LLM calls but reducing the risk of cross-attribute contamination. To further ensure robustness in our evaluation, we perform few-shot evaluation using human annotated in-context examples
on the same attributes. The actual prompt templates are provided in Table~\ref{tab:evaluation_prompts}. 

\textbf{Baselines \& Methods.}
We aim to augment LLMs with temporally relevant information for answering entity-centric questions and compare various augmentation strategies. Unlike existing RAG methods, which retrieve passages from large generic data indices, we employ strong baselines that retrieve from an entity-centric datastore (detailed in Section~\ref{sec:temporal_datastore}). We evaluate the following methods within the RAG framework to address our temporal QA task:
\begin{enumerate*}[label=(\roman*)]
    \item \textbf{NewsRAG:} Retrieves top-$k$ temporally indexed news chunks $C$, used in constructing \textsc{Neon} (step 2 in each variant of Section~\ref{sec:neon}).%

    \item \textbf{WebRAG:} Uses Bing to retrieve top-$k$ web snippets, explicitly incorporating timestamps to access diverse and dynamic sources.

    \item \textbf{\textsc{Neon($\mathcal{M}_1$)}:} Employs the process in Section~\ref{sec:neon1}; this variant includes 109,246 entries.

    \item \textbf{\textsc{Neon($\mathcal{M}_2$)}:} Employs the process in Section~\ref{sec:neon2}; this variant includes 455,680 entries.
\end{enumerate*}

Each method
uses the reformulated user query for retrieval, augments the top-$k$ retrieved information directly into the prompt, and employs GPT-4o for response generation. Other openIE-based methods are unsuitable for our purposes as they explicitly model knowledge tuples, which we aim to avoid. Comparing the performance across these methods
reveals how different forms of real-time information impact helpfulness, relevance, and faithfulness in addressing the temporal 
demands of our QA task.

\begin{table}[t!]
    \centering
    \begin{tabular}{p{0.1cm}lcccc}
    \toprule
  & \textbf{Method} & \textbf{Helpful} & \textbf{Relevant} & \textbf{Faithful} & \textbf{Avg.} \\ 
  \toprule
  \multirow{3}{*}{\small\rotatebox{90}{Temporal}} & NewsRAG & 1.03 & 0.83 & 0.83 & 0.9 \\  
   
  & \textsc{Neon($\mathcal{M}_1$)} & 1.03 & 0.82 & 0.86 & 0.9 \\
  & \textsc{Neon($\mathcal{M}_2$)} &  1.04 & 0.79 & 0.87 & 0.9 \\
  \midrule
  \multirow{3}{*}{\small\rotatebox{90}{Generic}} & NewsRAG & 0.9 & \textbf{1.24} & 1.08 & 1.07 \\
  & \textsc{Neon($\mathcal{M}_1$)} & \textbf{1.38} & 1.12 & \textbf{1.18} & \textbf{1.23} \\ 
  & \textsc{Neon($\mathcal{M}_2$)} & 1.37 & 1.03 & 1.17 & 1.19 \\ 
  \midrule
  & WebRAG & \colorbox[gray]{0.9}{1.54} & \colorbox[gray]{0.9}{1.29} & \colorbox[gray]{0.9}{1.22} & \colorbox[gray]{0.9}{1.35} \\
\bottomrule
    \end{tabular}
    \caption{Performance with zero-shot evaluation prompts.}
    \vspace{-2mm}
    \label{tab:results_zero_shot}
    \vspace{-4mm}
\end{table}

\begin{table}[t!]
    \centering
    \begin{tabular}{p{0.1cm}lcccc}
    \toprule
  & \textbf{Method} & \textbf{Helpful} & \textbf{Relevant} & \textbf{Faithful} & \textbf{Avg.} \\ 
  \toprule
  \multirow{3}{*}{\small\rotatebox{90}{Temporal}}
  & NewsRAG & 1.53 & 1.45 & 1.46 & 1.48 \\   
  & \textsc{Neon($\mathcal{M}_1$)} & 1.53 & 1.50 & \textbf{1.52} & \textbf{1.52} \\
  & \textsc{Neon($\mathcal{M}_2$)} & 1.55 & \textbf{1.54} & \colorbox[gray]{0.9}{1.56} & \colorbox[gray]{0.9}{1.55} \\
  \midrule
  \multirow{3}{*}{\small\rotatebox{90}{Generic}} 
  & NewsRAG & \colorbox[gray]{0.9}{1.59} & 1.41 &  1.05 & 1.35 \\
  & \textsc{Neon($\mathcal{M}_1$)} & 1.42 & 1.34 & 1.05 & 1.27 \\ 
  & \textsc{Neon($\mathcal{M}_2$)} & 1.45 & 1.35 & 1.09 & 1.30 \\ 
  \midrule
  & WebRAG & \textbf{1.57} & \colorbox[gray]{0.9}{1.57} & 1.41 & \textbf{1.52} \\
\bottomrule
    \end{tabular}
    \caption{Performance using few-shot learning and reasoning based evaluation prompts.}
    \label{tab:results_few_shot}
    \vspace{-4mm}
\end{table}

\begin{figure*}[t!]
     \centering
     \includegraphics[width=0.95\linewidth]{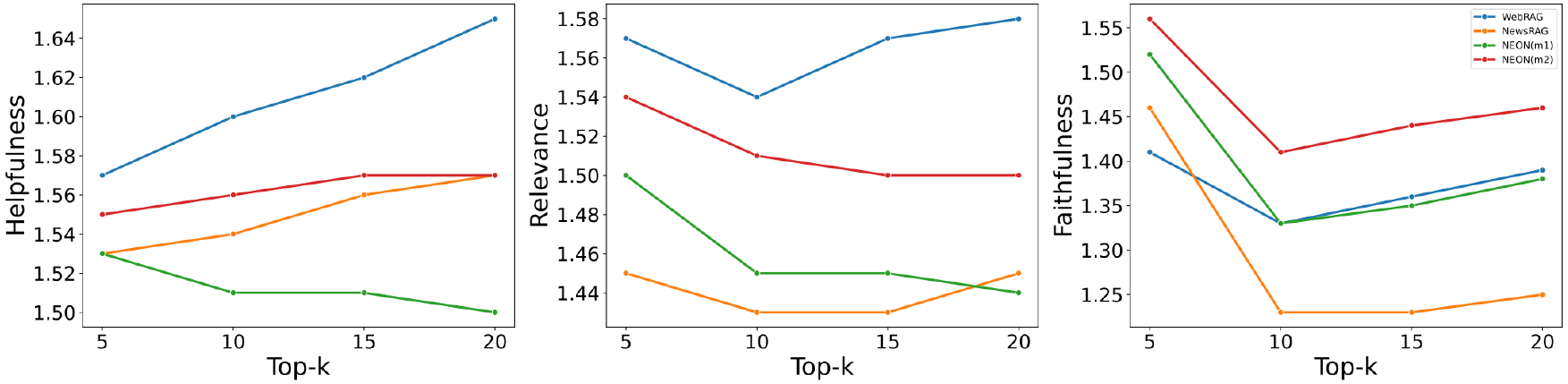}
     \Description{top-k with icl performance}
     \vspace{-3mm}
     \caption{Performance comparison with varying top-$k$ parameter, temporal retrieval and few-shot prompts.}
     \label{fig:varying_topk_performance_icl}
     \vspace{-3mm}
\end{figure*}

\section{Results}\label{sec:results}
Tables~\ref{tab:results_zero_shot} and~\ref{tab:results_few_shot} presents the main results, comparing system performance across \textit{temporal} and \textit{generic} retrieval strategies using a 3-point Likert scale
across the evaluation attributes and an average across the 3 scores. WebRAG is invariant to retrieval strategy since Bing %
combines aspects of both temporal and generic retrieval. Best scores are highlighted, while second-best scores are boldfaced.%

Beginning with Table~\ref{tab:results_zero_shot} we notice that in zero-shot setting WebRAG achieves the highest performance across all three attributes; this is unsurprising considering that it is achieved by the help of a full-fledged web-scale search engine. Meanwhile, \textsc{Neon} variants consistently outperform NewsRAG on both retrieval strategies for the helpfulness and faithfulness attributes, demonstrating that temporally relevant \textsc{Neon} KB entries improve generation reliability. Between \textsc{Neon} variants, both perform similarly on helpfulness and faithfulness, but \textsc{Neon($\mathcal{M}_1$)} achieves a higher relevance scores.

Few-shot learning with in-context examples and chain-of-thought prompts have been shown as being more effective for downstream tasks using LLMs~\cite{cot}. We apply this strategy and present the results in Table~\ref{tab:results_few_shot}. %
Unlike zero-shot evaluation, few-shot results show that temporal retrieval consistently outperforms generic retrieval across all methods. Additionally, much more significant gains are observed with \textsc{Neon} and NewsRAG methods. The results rival those of WebRAG in many instances, and \textsc{Neon($\mathcal{M}_2$)} achieves the highest overall average score across metrics. Moreover, each method shows specific strengths: NewsRAG produces the most helpful generations, WebRAG excels at relevance due to its broad and diverse sources, while \textsc{Neon} delivers the most reliable outputs, leveraging its structured entity-focused KB design. \textsc{Neon($\mathcal{M}_2$)} specifically benefits from a larger KB (4x the size of $\mathcal{M}_1$) and captures diverse entity-interactions by explicitly using subject and object mentions employed in the KB construction.

Overall, these results demonstrate that \textsc{Neon}---augmented with in-context examples---is capable of rivaling the performance of a live web-scale search engine for temporal entity-centric QA. Additionally, the individual strengths of the different approaches suggest a hybrid system that can leverage them jointly for even better performance; we leave such explorations to future work.

\textbf{Varying Context Length.}
The news chunks in NewsRAG typically contain 5-6 sentences, WebRAG uses 2-3 sentence web snippets, and \textsc{Neon} variants are single-sentence entries. To ensure comparable prompt context lengths across the methods, the no. of top-k documents is method-specific. Therefore, in Tables ~\ref{tab:results_zero_shot} and ~\ref{tab:results_few_shot}, NewsRAG and WebRAG each retrieve the top-5 documents, while \textsc{Neon} variants use top-10 documents. This creates a trade-off between detail and diversity, each of which may benefit a certain kind of query or response. To investigate this, we compare and present the performance with varying $k$, as illustrated in Figure~\ref{fig:varying_topk_performance_icl}. The order in performance for the studied models is relatively stable with the increase of $k$, with {WebRAG} leading \textsc{Neon} on helpfulness and relevance, most likely because of the capabilities of Bing to operate over a much larger search space. Meanwhile \textsc{Neon} remains the most faithful, but we notice a substantial drop for all models when doubling of supporting passages from 5 to 10, which appears to be more acute when passages are longer (NewsRAG and WebRAG).

\textbf{Human Assessment.}
To better understand and validate the automated evaluation results, we conducted a human assessment focusing on our 3 evaluation attributes. Annotators rated 100 randomly sampled responses across all methods using a 3-point Likert scale. To ensure consistency, annotators were provided with detailed descriptions of each attribute, method-specific supporting documents included in the LLM prompt for temporal QA, and examples illustrating each attribute-scale combination. External searches were disallowed, and ratings were based solely on the reformulated query, supporting documents, and LLM-generated responses.

The Pearson correlation between human assessments and few-shot prompt-based automated evaluations showed weak positive correlations: 0.24 (helpfulness), 0.12 (relevance), and 0.19 (faithfulness). Inter-annotator reliability, measured on the overlapping 10\% of the responses from 3 annotators using Krippendorff's alpha, indicated strong agreement with scores of 0.71 (helpfulness), 0.77 (relevance), and 0.61 (faithfulness).

A drill-down analysis of the ratings revealed that automated and human scores matched approximately 50\% of the time. Discrepancies in the remaining cases often stemmed from mismatches in timestamps across the query, supporting passages, and LLM responses. Automated evaluations consistently assigned a score of 0 in these cases, whereas human annotators used common-sense reasoning and contextual understanding of the dates to assign scores of 1 or 2. Addressing this limitation in automated evaluation is a goal for future work. In particular, such mismatches were less frequent in samples from \textsc{Neon($\mathcal{M}_2$)}.

Interestingly, the ratings for helpfulness and relevance, both human and automated, exhibited a high correlation (>0.8). This is likely because a highly relevant response---one that accurately reflects the query's specific time frame---almost always leads to a helpful response. The reverse however is not always true, as a helpful response may not strictly adhere to temporal relevance.

\begin{table*}
\resizebox{\textwidth}{!}
{
\begin{tabular}{p{0.06cm}p{1.3cm}p{0.9cm}p{1.2cm}p{5.7cm}p{6cm}}
\hline
\toprule
& Query & Date & Method & Retrieved Passages & Temporal QA Response\\
\toprule
\multirow{6}{*}{\rotatebox{90}{ex. 1}}
  & Sherry Rehman news & Oct 18, 2023 & WebRAG & \small{[1] Waheeda Rehman Biography:- ... born in Chengalpet ... [2] Sherry Rehman, parlementaire du Pakistan ... [3] For a sweeter white ... substituting sherry for white wine ... [4] ... PX Sherry Cask Matured limited whisky ... [5] Waheeda Rehman (born 3 February 1938) is an Indian ...}  & \small{On October 18, 2023, Sherry Rehman, a prominent member of the Pakistan People's Party, is in the news due to her involvement in a legal case concern ... This development highlights the ongoing and contentious debate surrounding blasphemy laws in Pakistan and the risks faced by those who seek to reform them.} \\
\hline

\multirow{6}{*}{\rotatebox{90}{ex. 2}}
  & Olaf Scholz eye patch & Sept 4, 2023 & WebRAG & \small{[1] Chancellor `Pirate Olaf' Scholz injured jogging ... [2] ... Olaf Scholz. Can someone Photoshop an eye patch ... [3] Après avoir chuté lors d’un footing ce samedi 2 septembre 2023, le chancelier allemand Olaf Scholz s’est ... [4] ... Olaf Scholz Pirat is not real ... [5] Olaf ... ready for the great pirate era ... } & \small{On September 2, 2023, German Chancellor Olaf Scholz sustained minor facial injuries while jogging ... he appeared in public on Monday, September 4, 2023, wearing an eye patch, which has led to a wave of humorous and supportive reactions online ... playful nickname ``Pirate Olaf'' among social media users ...} \\
\toprule
& Query & Date & Method & Temporal QA Response & LLM Rating\\
\toprule
\multirow{6}{*}{\rotatebox{90}{ex. 3}}
  & Sam Altman Microsoft & Nov 22, 2023 & \textsc{Neon($\mathcal{M}_2$)} & \small{As of November 22, 2023, Sam Altman has been significantly involved with Microsoft ... OpenAI, which Altman co-founded, is 49\% owned by Microsoft ... a move supported by Nadella ... In summary, as of late November 2023, Sam Altman continues to play a pivotal role in the collaboration ...}& \small{\{``reason'': ``The response provides a detailed and relevant answer to the user's query about Sam Altman's involvement with Microsoft as of November 22, 2023. It includes information about his role, the financial ... The response is accurate, informative, and directly addresses the user's question''. ``rating'': 5\}}\\
\bottomrule
\end{tabular}
}
\caption{Examples of queries and LLM responses based on retrieved passages and a corresponding LLM rating (ex. 3)}
\label{tab:qualitative_examples}
\vspace{-5mm}
\end{table*}

\section{Qualitative Analysis}\label{sec:qualitative_analysis}
We perform a qualitative analysis to gain insights into the successes and failures of each method. By performing such an inspection, we identified patterns that either support or hinder the generation of temporally relevant responses. Table~\ref{tab:qualitative_examples} highlights key cases.

\noindent \textit{Entity-centric temporal RAG enhances reliability.} As emphasized earlier, unlike traditional RAG techniques, our approach leverages an entity-centric temporal datastore, which improves retrieval accuracy. This advantage is evident in methods using news chunks (\textsc{Neon} variants and NewsRAG), as Bing search in WebRAG relies on surface form matches that often retrieve irrelevant snippets, particularly for long-tail entities with low search hit rates (ex.1 in Table~\ref{tab:qualitative_examples}). Consequently, the entity-centric design of \textsc{Neon} leads to more reliable responses.

\noindent \textit{WebRAG benefits from non-EN information and LLMs.} We noticed that Bing search sometimes retrieves noisy passages together with relevant passages that are in languages other than English, particularly for subject entities from non-English demographic regions. In such cases, the LLM is still able to generate temporally relevant responses in English based on the non-English passages (ex.2 in Table~\ref{tab:qualitative_examples}). This gives WebRAG an advantage over others, as the news streams we employed are exclusively in English.

\noindent \textit{Deviations from rating instructions.} We observed that LLMs do not always adhere to the provided evaluation instructions. Despite the prompt clearly outlining the criteria for each 3-point Likert rating, the model occasionally generated ratings outside the intended range along with reasons for such ratings. We noticed a response that was assigned a rating of 3 when the model described it as somewhat relevant and responses with ratings of 5 and 9 when they were deemed fully reliable and relevant (ex.3 in Table~\ref{tab:qualitative_examples}).

\noindent \textit{Response lengths correlate with Likert scores.} We analyzed whether longer temporal responses tend to receive higher Likert scores by examining the correlation between ratings and response length (measured in characters). Our analyses (see~Table~\ref{tab:length_correlation} in the Appendix for more details) reveal the following trends: (i) in case of helpfulness and relevance, both mean and median response lengths increase with higher ratings, particularly for NewsRAG, indicating that longer responses are more likely to receive higher scores. (ii) for faithfulness, while mean and median response lengths do not consistently increase with higher ratings, NewsRAG shows a clear positive correlation between longer responses and higher ratings. 

As NewsRAG uses the longest supporting information---5-6 sentence news chunks---these findings suggest that for robust automated evaluation, it is valuable to augment concise yet informative documents, similar to \textsc{Neon}, into LLM prompts to promote consistent response lengths. Notably, LLM-generated reasons for a rating of 2 frequently included the phrases ``the response provides detailed and relevant information'' for helpfulness and relevance, and ``the response accurately reflects the information'' for faithfulness.

\section{Related Work}

\textbf{Information Extraction.}
Extracting structured information from unstructured text has been a central focus of NLP research~\cite{han-etal-2020-data}. A key task in IE is relation extraction (RE), which identifies relationships between entities using named entity recognition, entity linking, and relational classification. Accurate relation identification, however, requires deep semantic understanding of context and is often constrained by the fixed relation types in traditional classifier-based RE methods~\cite{ma-etal-2023-dreeam}.

To address these limitations, openIE enables entities and relations to be extracted as surface forms or phrases~\cite{kolluru-etal-2020-openie6}. Both RE and openIE are vital for constructing and enriching knowledge graphs~\cite{10.1162/tacl_a_00456, 10.1016/j.eswa.2018.07.017}. Recent advances explore LLMs as implicit KBs~\cite{petroni-etal-2019-language}, leveraging their outputs for IE tasks via prompting\cite{alkhamissi2022reviewlanguagemodelsknowledge, hao-etal-2023-bertnet, cohen-etal-2023-crawling}. Our work leverages LLMs for openIE-style lexicalized extraction on news chunks, capturing interactions such as emerging events and associations between entities.

\textbf{KGs for Temporal QA.}  Temporal QA aims to generate answers for queries with time-specific constraints~\cite{saxena-etal-2021-question, mavromatis2022tempoqr}. KGs provide 
structured and explainable information that can complement LLMs in answering questions. However, these approaches have limitations in handling evolving entity-specific temporal queries and mostly focus on temporal reasoning tasks~\cite{jia2024faithful, saxena-etal-2021-question}. Our work builds on these existing methods by introducing a novel way to incorporate entity-interactions from news streams into the QA process, thereby addressing the temporal and entity-specific challenges.

\textbf{Retrieval Models.}
Recent research has focused on retrieval-augmented LLMs, which combine information from external sources with parametric models to improve the factuality of text generation~\cite{realm, NEURIPS2020_6b493230, asai2024selfrag}. But extensive study on entity-centric temporal RAG remain scarce. For instance, GraphRAG~\citep{edge2024localglobalgraphrag} builds entity KGs and generates summarizes for closely related entities to support query responses.  While effective at capturing global dependencies, these summaries can overlook nuanced, temporally evolving entity-interactions that are critical for reliable temporal QA.

\section{Conclusion}
\textsc{Neon} presents a novel approach to enhancing QA systems by leveraging entity-interactions extracted from news streams. By constructing an entity-focused time-stamped graph and integrating its information into LLM prompts, \textsc{Neon} addresses the limitations of existing QA methods in handling temporal and entity-specific queries. Our experimental results demonstrate the effectiveness of \textsc{Neon} in generating relevant and reliable answers, making it a valuable addition to the QA landscape.

Future work will focus on further improving the \textsc{Neon} system by incorporating more sophisticated entity-interaction extraction techniques, strengthening the evaluation by expanding the dataset to include more entities and queries, and exploring additional applications to this framework such as enterprise and personal data integration.

\bibliographystyle{ACM-Reference-Format}
\bibliography{ref}

\appendix

\setcounter{figure}{0}
\renewcommand{\thefigure}{A\arabic{figure}}

\begin{figure}[b!]
    
     \centering
     \fbox{\includegraphics[width=0.9\linewidth]{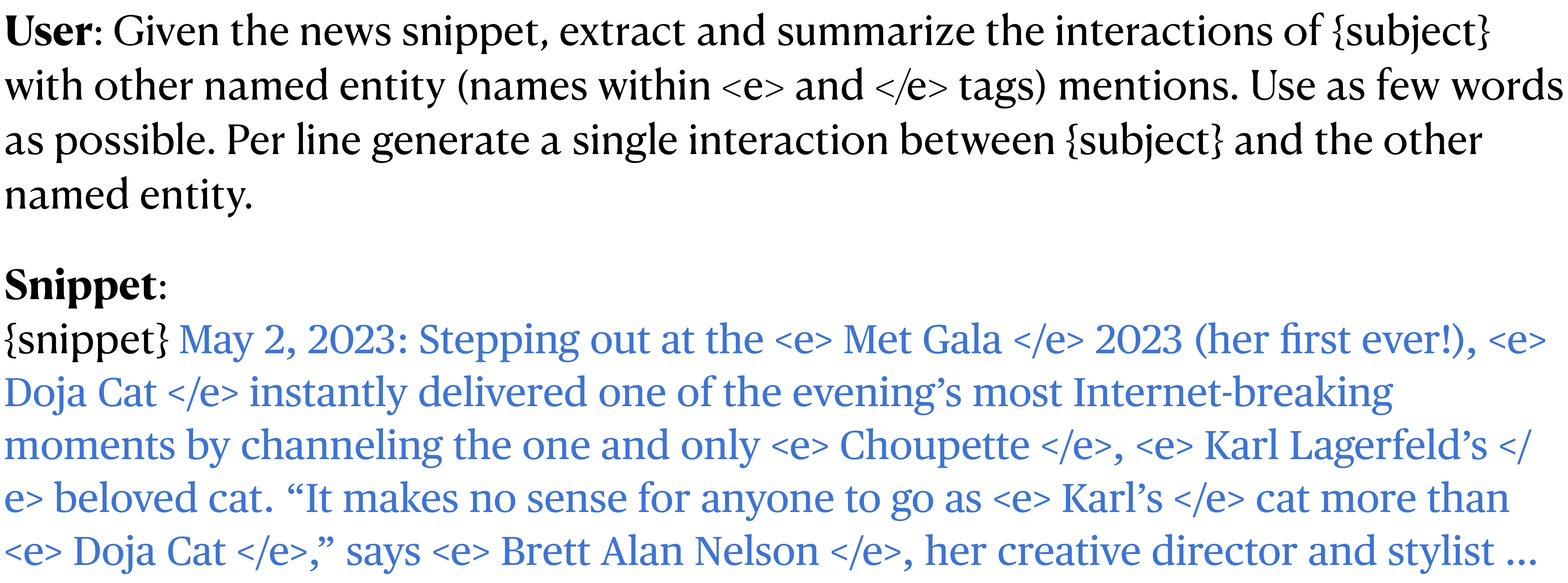}}
     \Description{Neon generation 1}
     \vspace{-3mm}
     \caption{Prompt for \textsc{Neon}($\mathcal{M}_1$) construction.}
     \label{fig:llm_prompt1}
     \vspace{-1mm}
\end{figure}

\begin{figure}[b]
     \centering
     \fbox{\includegraphics[width=0.9\linewidth]{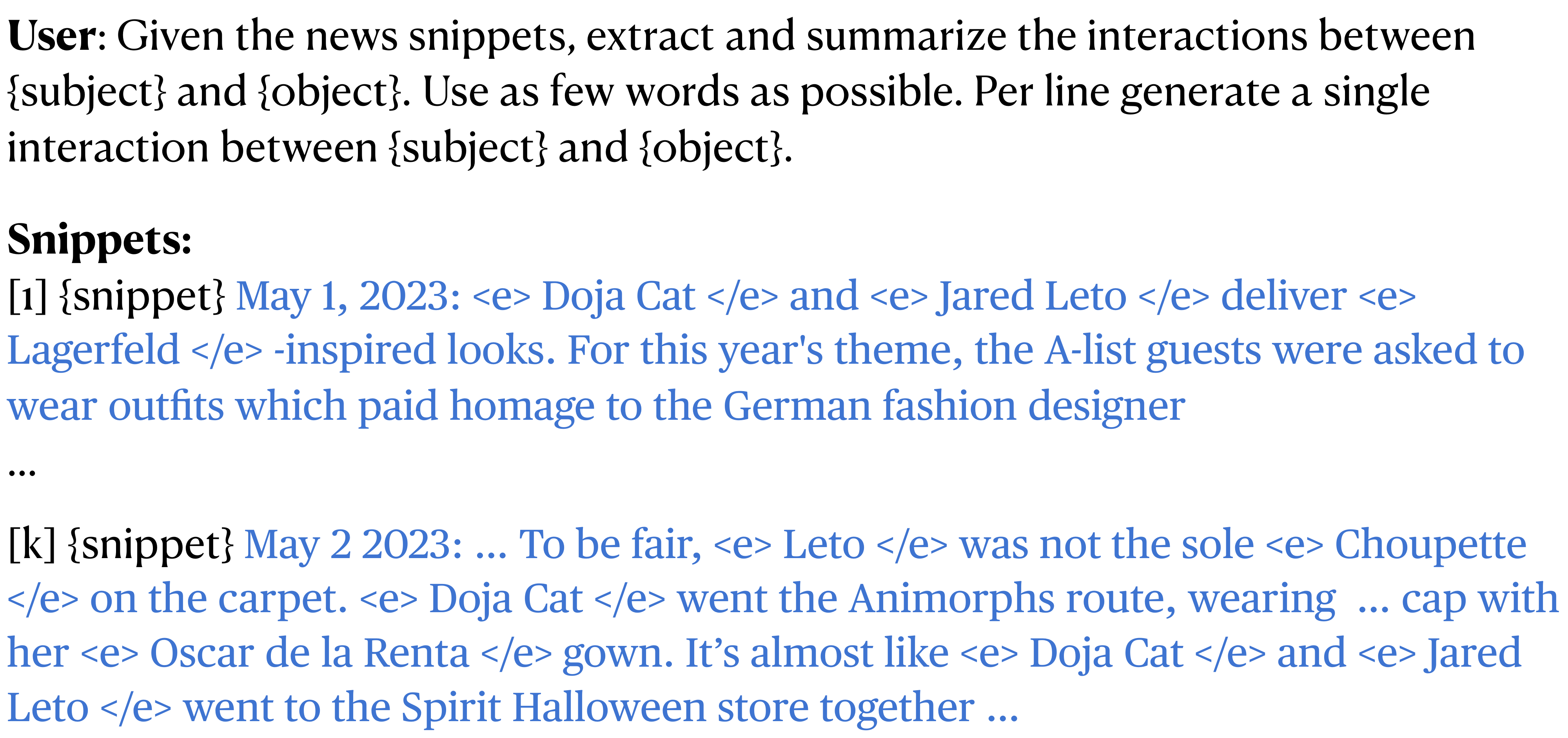}}
     \Description{Neon generation 2}
     \vspace{-3mm}
     \caption{Prompt for \textsc{Neon}($\mathcal{M}_2$) construction.}
     \label{fig:llm_prompt2}
     \vspace{-1mm}
\end{figure}

\setcounter{table}{2}
\renewcommand{\thetable}{A\arabic{table}}

\begin{table}[b]
\resizebox{0.9\columnwidth}{!}
{
\begin{tabular}{cc|cc|cc|cc}
\hline
 \multicolumn{2}{c}{\multirow{2}{*}{Score}}  & \multicolumn{2}{c}{Helpfulness} & \multicolumn{2}{c}{Relevance} & \multicolumn{2}{c}{Faithfulness} \\
 & & Mean & Median  & Mean & Median  & Mean & Median  \\
\hline
\multirow{3}{*}{\rotatebox{90}{\footnotesize{WebRAG}}} & 0  & 705.8 & 591.0  & 750.6 & 651.0  & 940.2 & 871.5 \\
 & 1      & 801.7 & 731.0  & 911.2 & 884.0  & 905.3 & 877.0  \\
 & 2      & 908.2 & 839.0  & 865.0 & 762.0  & 814.5 & 688.0  \\
 \hline
\multirow{3}{*}{\rotatebox{90}{\footnotesize\textsc{Neon($\mathcal{M}_1$)}}}  & 0      & 641.5 & 609.0   & 674.7 & 637.0  & 786.7 & 706.5   \\
 & 1      & 710.7 & 661.0   & 756.2 & 692.0   & 755.1 & 697.0  \\
 & 2      & 819.6 & 736.0   & 798.5 & 709.5   & 774.5 & 679.5  \\
 \hline
\multirow{3}{*}{\rotatebox{90}{\footnotesize\textsc{Neon($\mathcal{M}_2$)}}}  & 0      & 634.4 & 584.0    & 666.3 & 596.5  & 820.2 & 723.5  \\
 & 1      & 713.3 & 648.0   & 763.8 & 700.0   & 735.0 & 650.0   \\
 & 2      & 837.0 & 753.0   & 813.4 & 710.0   & 798.3 & 701.0   \\
 \hline
\multirow{3}{*}{\rotatebox{90}{\footnotesize{NewsRAG}}}  & 0      & 919.5 & 827.5  & 968.5 & 887.0  & 1155.0 & 1113.0  \\
 & 1      & 1042.6 & 992.5    & 1108.4 & 1088.0  & 1159.2 & 1117.5   \\
 & 2      & 1358.0 & 1342.0  & 1347.7 & 1333.5  & 1275.0 & 1247.5  \\
 \hline
\end{tabular}}
\caption{Response length statistics (number of chars)}
\label{tab:length_correlation}
\vspace{-3mm}
\end{table}

\begin{table}[]
    \centering
    \begin{tabular}{p{80mm}}
    \toprule
    \small{Helpfulness}\\
    \midrule
    \footnotesize{\#\# \textbf{Task Description} You are presented with a user query and an AI assistant's response. The query is focused on a specific entity, pertains to the news domain, and is date-stamped. Your task is to evaluate the AI assistant's response for its usefulness, using a 3-point Likert scale. The criteria for rating are detailed below.
    \#\# Helpfulness Criterion Rating 2: The response is very helpful and provides the information expected for the user query. It includes mentions of additional named entities (such as people, locations, events, etc.) beyond the primary entity in the query and aligns completely with user's intent. Rating 1: The response is somewhat helpful but fails to fully provide the information expected for the user's query. It can nevertheless serve to continue the conversation with the user or provides pointers to where the information can be found. Rating 0: The response is not helpful and provides no information for the query.
    \#\# Output format The output should be the following JSON format: \{"rating": <numerical\_rating>, "reason": <short\_reasoning>\}, mentioning the numerical rating, as well as a short and concise reasoning for the helpfulness rating.
    \#\# Examples \{in-context examples\}
    \#\# Input to be rated  User query: \{question\} AI assistant’s response: \{response\}}\\
    \midrule
        \small{Relevance}\\
    \midrule
    \footnotesize{\#\# \textbf{Task Description} You are presented with a user query and an AI assistant's response. The query is focused on a specific entity, pertains to the news domain, and is date-stamped. Your task is to evaluate the AI assistant's response for its relevance, using a 3-point Likert scale. The criteria for rating are detailed below.
\#\# Relevance Criterion Rating 2: The response is completely relevant with accurate details and provides the information for the query date. Rating 1: The response contains a mix of relevant and irrelevant details. The response contains some relevant information upto the specified date, and is more or less aligned with the user's intent. Rating 0: The response is incorrect and provides no information for the query date.
\#\# Output format The output should be the following JSON format: \{"rating": <numerical\_rating>, "reason": <short\_reasoning>\}, mentioning the numerical rating, as well as a short and concise reasoning for the helpfulness rating.
\#\# Examples \{in-context examples\}
\#\# Input to be rated  User query: \{question\} AI assistant’s response: \{response\}}\\
    \midrule
        \small{Faithfulness}\\
    \midrule
    \footnotesize{
    \#\# \textbf{Task Description} You are presented with a user query and an AI assistant's response. The query is focused on a specific entity, pertains to the news domain, and is date-stamped. Your task is to evaluate the AI assistant's response for its reliability, using a 3-point Likert scale. The criteria for rating are detailed below.
\#\# Faithfulness Criterion Rating 2: The response is perfectly reliable and grounded based on the supporting passages given below. All the information from the supporting passages is used in the response to answer the user query. Rating 1: The response partially uses the supporting passages given below but has additional information which may be incorrect or unreliable. Rating 0: The response is completely unreliable and does not depend on the supporting passages.
\#\# Output format The output should be the following JSON format: \{"rating": <numerical\_rating>, "reason": <short\_reasoning>\}, mentioning the numerical rating, as well as a short and concise reasoning for the helpfulness rating.
\#\# Examples \{in-context examples\}
\#\# Input to be rated  User query: \{question\} Supporting passages: \{passages\} AI assistant’s response: \{response\}}\\
    \bottomrule
    \end{tabular}
    \caption{Prompt templates for automated evaluation using few-shot learning.}
    \label{tab:evaluation_prompts}
\end{table}

\end{document}